\documentclass{article}

\usepackage[preprint]{corl_2026}    

\makeatletter
\renewcommand{\@noticestring}{Submitted to the 4th Workshop on Learning Effective Abstractions for Planning (LEAP) at CoRL 2026. Do not distribute.}
\makeatother

\usepackage{amsmath,amssymb}
\usepackage{graphicx}
\usepackage{wrapfig}
\usepackage{booktabs}
\usepackage[draft]{labreview}       

\title{CausalDreamer: Learning Predictive World Models with Latent Disentanglement}

\author{
  Prince Jha$^{1}$ \qquad Nils Lukas$^{1}$ \qquad Kun Zhang$^{1,2}$ \qquad Salem Lahlou$^{1}$ \\[4pt]
  \normalfont $^{1}$Mohamed bin Zayed University of Artificial Intelligence (MBZUAI) \\
  \normalfont $^{2}$Carnegie Mellon University \\
}

\begin{document}
\maketitle

\begin{abstract}
World models for control must capture which aspects of the environment respond to the agent's actions and which are relevant to reward. Generative world models such as Dreamer 4 consist of a video tokenizer, which encodes each frame into a latent, and a dynamics model, which is pretrained to predict future latents from past latents and actions. Yet the tokenizer is trained with a reconstruction objective, without action or reward supervision, so its latent provides no explicit mechanism to separate controllable, uncontrollable, reward-relevant, and reward-irrelevant information. We propose \textit{CausalDreamer}, which keeps the tokenizer frozen and re-encodes its latent into a factored representation of four groups along two axes: controllability, where only the two controllable groups receive the action, and reward relevance, learned by predicting the reward from the two reward-relevant groups. The pretrained dynamics model is then fine-tuned to predict the factored representation. We evaluate \textit{CausalDreamer} and the pretrained world model it starts from with model-predictive planning on 20 MMBench2 tasks: 10 clean tasks seen during training and 10 unseen tasks, of which 6 are manipulated variants of clean tasks with a changed background, object, or maze layout, and 4 are new environments. We normalize returns so that a policy taking uniformly random actions scores 0 and an expert scores 1. \textit{CausalDreamer} achieves a 14\% higher normalized score than the pretrained world model on the clean tasks (0.199 vs.\ 0.175) and a 25\% higher score on the manipulated variants (0.307 vs.\ 0.246), while neither model scores meaningfully above the random policy in the new environments. Additionally, our analysis shows that the factored representation separates reward-irrelevant changes, such as a changed background, from its reward-relevant groups.
\end{abstract}


\section{Introduction}
\label{sec:introduction}

World models let an agent plan by predicting the consequences of candidate actions in a learned latent space, rather than by trial and error in the environment~\citep{hafner2023dreamerv3, hansen2024tdmpc2, hafner2025dreamer4}.
For such planning to be reliable, the latent must expose the structure that decisions depend on, namely which parts of the scene respond to the agent's actions and which parts determine reward.
A latent that mixes the agent's own effect with background motion, or task-relevant objects with irrelevant texture, forces the planner to compare imagined futures that differ in ways unrelated to the task~\citep{wang2022denoised, liu2023ifactor}.

Scalable world models now render realistic, action-conditioned video by pairing a video tokenizer with a diffusion or flow-matching dynamics model trained on hundreds of hours of multi-task data~\citep{hafner2025dreamer4, hansen2026mmbench2}.
However, their tokenizer learns its latent from a reconstruction objective, without access to actions or rewards, so the latent does not explicitly distinguish controllable from uncontrollable content, or reward-relevant from reward-irrelevant content.
The dynamics model must discover these distinctions inside one entangled latent, and in practice it can predict frames that are visually plausible but ignore the action actually taken~\citep{hansen2026mmbench2}.
Methods that do learn such factorized latents rely on variational recurrent state-space models or non-generative predictors trained from scratch, and impose the factorization through per-factor KL terms or mutual-information constraints~\citep{fu2021learning, wang2022denoised, liu2023ifactor}.
These mechanisms have no counterpart in a tokenizer-based generative world model, which leaves open how controllability and reward structure can be built into such a model while its tokenizer remains frozen.

\begin{wrapfigure}{r}{0.5\linewidth}
  \vspace{-10pt}
  \centering
  \includegraphics[width=\linewidth]{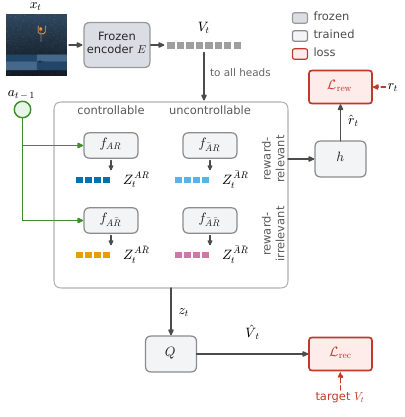}
  \caption{\textbf{Learning the factored representation.} A frozen encoder maps each frame to the tokenizer representation. Four projection heads re-encode it into four groups, organized by controllability and reward relevance; only the two controllable heads receive the action. A reward predictor reads only the two reward-relevant groups, and a residual decoder reconstructs the tokenizer representation from all four groups. Grey components are frozen, and the red boxes are the two training losses.}
  \label{fig:method}
  \vspace{-10pt}
\end{wrapfigure}

Accordingly, we propose \emph{CausalDreamer}, which places a factored representation between the frozen tokenizer of a pretrained generative world model and its dynamics model.
Each frame's latent is re-encoded into four groups, organized by whether they are controllable and whether they are relevant to reward.
Only the two controllable groups receive the action, so the routing of action information is an architectural choice rather than a learned one.
Reward relevance is encouraged by predicting the reward from the two reward-relevant groups only.
The four groups jointly reconstruct the tokenizer latent, so the frozen decoder still renders them, and together they keep the latent's shape, so the pretrained dynamics model is fine-tuned to predict the factored representation with its original architecture and losses.
CausalDreamer therefore adds this structure to an existing generative world model after pretraining, with the tokenizer frozen and without a variational posterior or mutual-information estimator.

We build CausalDreamer on the 350M-parameter MMBench2 world model and compare it with the pretrained model it is fine-tuned from, using closed-loop planning with the cross-entropy method.
The evaluation covers 10 clean tasks seen during training and 10 unseen tasks: 6 manipulated variants of clean tasks, in which the background, an object, or the maze layout differs, and 4 new environments.
Returns are normalized between a policy that takes uniformly random actions (0) and an expert (1).
On the clean tasks, CausalDreamer raises the normalized score from 0.175 to 0.199, a 14\% improvement, and on the manipulated variants from 0.246 to 0.307, a 25\% improvement; in the new environments, neither model plans meaningfully better than the random policy.
Finally, when only the background of a scene changes, most of the resulting change in the factored representation stays out of its reward-relevant groups, which shows that the factored representation separates reward-irrelevant changes from the content used to predict reward.

In summary, our contributions are threefold.
\textbf{(i)}~We introduce \textit{CausalDreamer}, which re-encodes the latent of a frozen video tokenizer into a \textit{factored representation} along controllability and reward relevance, and fine-tunes the pretrained dynamics model on it.
\textbf{(ii)}~We show that planning with CausalDreamer reaches a \textit{higher normalized score} than the pretrained model on the clean tasks of MMBench2 and on manipulated variants of these tasks not seen during training.
\textbf{(iii)}~We show, by changing only the background of a scene, that the factored representation \textit{keeps reward-irrelevant changes largely out of its reward-relevant groups}.

\section{Background}
\label{sec:background}

We build on the world model of MMBench2~\citep{hansen2026mmbench2}, a 350M-parameter model that follows the architecture and training recipe of Dreamer 4~\citep{hafner2025dreamer4}: a video tokenizer is trained first, and a dynamics model is then trained on the representations of the frozen tokenizer.

\textbf{Tokenizer.}
The tokenizer is a transformer encoder--decoder.
The encoder $E$ splits each $224 \times 224$ frame into 256 patches and prepends $L = 64$ learned latent tokens, which attend to all patches while the patches attend only to each other; the latent tokens are projected to $D = 64$ channels and bounded by a $\tanh$, giving $V_t \in [-1,1]^{L \times D}$.
The decoder $G$ reconstructs the frame from $V_t$ alone.
The tokenizer is trained with mean squared error and LPIPS~\citep{zhang2018unreasonable}, $\mathcal{L}_{\mathrm{tok}} = \mathcal{L}_{\mathrm{MSE}} + 0.2\,\mathcal{L}_{\mathrm{LPIPS}}$, in a masked-autoencoding setup~\citep{he2022masked} in which a fraction $p \sim U(0, 0.9)$ of the input patches of each frame is replaced by a learned mask token.

\textbf{Dynamics Model.}
The dynamics model $f_\theta$ is a block-causal transformer over the frames of a clip: tokens of frame $t$ attend to each other and to the tokens of earlier frames.
For each frame, its input consists of an action token, a token encoding the signal level and step size of the shortcut objective below, 32 spatial tokens obtained by packing pairs of the $L$ latent vectors of $V_t$, and 4 learned register tokens.
For control, agent tokens that receive an embedding of the task description are added to each frame~\citep{hafner2025dreamer4}; they attend to all other tokens, while no other token attends to them, so that the task does not influence the predicted frames.
A reward head and a behavior-cloning head read the agent tokens and predict the rewards and actions of the next $M = 8$ steps, trained with a two-hot cross-entropy over symlog-transformed reward bins ($\mathcal{L}_{\mathrm{rew}}^{\mathrm{dyn}}$) and a mean squared error on the actions ($\mathcal{L}_{\mathrm{bc}}$)~\citep{hansen2026mmbench2}.

\textbf{Shortcut Forcing.}
The dynamics model is trained with the shortcut forcing objective of Dreamer 4, which combines diffusion forcing~\citep{chen2024diffusion}, where every frame of a clip receives its own noise level, with shortcut models~\citep{frans2025one}, where the network is also conditioned on the size of the denoising step.
Each frame is corrupted to $\tilde{V}_t = (1 - \tau_t)\,\epsilon + \tau_t V_t$ with $\epsilon \sim \mathcal{N}(0, I)$, where the signal level $\tau_t \in [0,1]$ runs from pure noise ($\tau_t = 0$) to the clean frame ($\tau_t = 1$), and the network predicts the clean representations from the corrupted clip, the actions, the signal levels and the step sizes $d_t$,
\begin{equation}
\hat{V} = f_\theta\big(\tilde{V}, \tau, d, a\big).
\label{eq:bg_pred}
\end{equation}
At the finest step size $d_{\min}$, the loss is the squared error to the clean representation; for larger steps, the target is obtained by bootstrapping two half steps of the network itself, $b' = \big(f_\theta(\tilde{V}, \tau, \tfrac{d}{2}, a) - \tilde{V}\big)/(1 - \tau)$, $V' = \tilde{V} + \tfrac{d}{2}\, b'$ and $b'' = \big(f_\theta(V', \tau + \tfrac{d}{2}, \tfrac{d}{2}, a) - V'\big)/(1 - \tau - \tfrac{d}{2})$, per frame,
\begin{equation}
\mathcal{L}_{\mathrm{shortcut}} = \mathbb{E}\Big[\, w(\tau) \cdot
\begin{cases}
\lVert \hat{V} - V \rVert_2^2, & d = d_{\min},\\[2pt]
(1 - \tau)^2 \,\big\lVert (\hat{V} - \tilde{V})/(1 - \tau) - \mathrm{sg}\big((b' + b'')/2\big) \big\rVert_2^2, & \text{otherwise},
\end{cases}
\Big]
\label{eq:shortcut}
\end{equation}
where $\mathrm{sg}$ stops gradients and the ramp weight $w(\tau) = 0.9\,\tau + 0.1$ emphasizes the less noisy signal levels.
At inference, each frame is generated from noise in $K = 4$ steps of size $d = 1/4$, conditioned on the earlier, already generated frames, which are slightly corrupted to signal level $\tau_{\mathrm{ctx}} = 0.1$.
The full dynamics objective is $\mathcal{L}_{\mathrm{dyn}} = \mathcal{L}_{\mathrm{shortcut}} + \mathcal{L}_{\mathrm{rew}}^{\mathrm{dyn}} + \mathcal{L}_{\mathrm{bc}}$.

\section{Method}
\label{sec:method}

CausalDreamer adds a factored representation to the pretrained world model of Section~\ref{sec:background} in two stages, keeping the tokenizer frozen throughout.
We first define the factored representation and the role of each of its groups (Section~\ref{sec:method_setting}).
We then describe how CausalDreamer learns this representation from the tokenizer representation (Figure~\ref{fig:method}) and how the pretrained dynamics model is fine-tuned to predict it (Section~\ref{sec:method_causaldreamer}).

\subsection{Problem Setting}
\label{sec:method_setting}

We consider an agent interacting with an environment, modeled as a partially observed Markov decision process with latent states $s_t$, actions $a_t$, observations $x_t$ and rewards $r_t$.
At each step, the environment moves from $s_{t-1}$ to $s_t \sim P(\cdot \mid s_{t-1}, a_{t-1})$ under the action $a_{t-1}$ taken by the agent, emits an image observation $x_t \in \mathbb{R}^{H \times W \times C}$ of the new state, and returns the reward $r_t = R(s_{t-1}, a_{t-1})$ of this transition.
The data available for learning consist of clips of $T$ consecutive steps $(x_t, a_{t-1}, r_t)_{t=1}^{T}$.

\textbf{Tokenizer Representation.}
Each frame is represented by the frozen tokenizer of Section~\ref{sec:background}, which maps it, together with the earlier frames of its clip, to
\begin{equation}
V_t = E(x_{\le t}) \in [-1, 1]^{L \times D}.
\label{eq:tokenizer}
\end{equation}
Since the tokenizer is trained without action or reward supervision, $V_t$ provides no explicit mechanism to separate controllable, uncontrollable, reward-relevant, and reward-irrelevant information.

\textbf{Factored Representation.}
Given the tokenizer representation $V_t$ of a frame and the action $a_{t-1}$, we aim to learn a factored representation $z_t$ that supports reconstruction of the tokenizer representation and prediction of the reward.
The factored representation consists of four groups $Z^k_t \in \mathbb{R}^{(L/4) \times D}$, indexed by $k \in \mathcal{K} = \{AR, \bar{A}R, A\bar{R}, \bar{A}\bar{R}\}$, where each label names the combination of controllability ($A$ or $\bar{A}$) and reward relevance ($R$ or $\bar{R}$) the group is designed to capture,
\begin{equation}
\big(V_t, a_{t-1}\big) \;\mapsto\; z_t = \big(Z^{AR}_t, Z^{\bar{A}R}_t, Z^{A\bar{R}}_t, Z^{\bar{A}\bar{R}}_t\big) \;\mapsto\; \big(\hat{V}_t, \hat{r}_t\big).
\label{eq:objective}
\end{equation}
The \emph{controllable} groups $\mathcal{K}_A = \{AR, A\bar{R}\}$ depend on $V_t$ and $a_{t-1}$, and the uncontrollable groups depend on $V_t$ alone,
\begin{equation}
Z^k_t =
\begin{cases}
f_k(V_t, a_{t-1}), & k \in \mathcal{K}_A,\\
f_k(V_t), & k \notin \mathcal{K}_A.
\end{cases}
\label{eq:controllable}
\end{equation}
The \emph{reward-relevant} groups $\mathcal{K}_R = \{AR, \bar{A}R\}$ are used to predict the reward,
\begin{equation}
\hat{r}_t = h\big(Z^{AR}_t, Z^{\bar{A}R}_t\big).
\label{eq:reward}
\end{equation}

\subsection{CausalDreamer}
\label{sec:method_causaldreamer}

\textbf{Action-Conditioned Projections.}
Each projection $f_k$ in Eq.~\eqref{eq:controllable} is a multilayer perceptron that reads the whole tokenizer representation flattened into one vector $v_t = \mathrm{vec}(V_t) \in \mathbb{R}^{LD}$, with the action concatenated, $[\,v_t;\, a_{t-1}\,]$, for $k \in \mathcal{K}_A$, where $a_{t-1}$ is set to zero for the first frame of a clip.
Since every head reads all of $V_t$, each group is a learned re-encoding of the whole frame rather than a subset of the tokenizer's latent vectors.
Each head outputs a vector of size $LD/4$, which is reshaped to $Z^k_t \in \mathbb{R}^{(L/4) \times D}$, and the four groups are stacked in the order of Eq.~\eqref{eq:objective} to form $z_t \in \mathbb{R}^{L \times D}$.

\textbf{Learning Objective.}
Figure~\ref{fig:method} illustrates how the factored representation is learned.
A residual decoder $Q(z) = z + q(z)$, where $q$ is a multilayer perceptron whose last layer is initialized to zero, maps the concatenated groups back to the tokenizer representation, and a reward predictor $h$ reads only the reward-relevant groups.
With the tokenizer frozen, the heads, $q$ and $h$ are trained per frame with
\begin{equation}
\mathcal{L}_{\mathrm{fac}}
= \underbrace{\mathbb{E}\Big[\big\lVert Q(z_t) - v_t \big\rVert_2^2\Big]}_{\mathcal{L}_{\mathrm{rec}}}
+ \lambda \, \underbrace{\mathbb{E}\Big[\mathrm{CE}\big(h(Z^{AR}_t, Z^{\bar{A}R}_t),\, r_t\big)\Big]}_{\mathcal{L}_{\mathrm{rew}}},
\label{eq:fac}
\end{equation}
where $\mathrm{CE}$ is a two-hot cross-entropy over symlog-transformed reward bins, the first frame of each clip has no reward target, and $\lambda$ weights the reward term.
The reconstruction term encourages the four groups together to retain the information in $V_t$, and lets the frozen decoder render a factored representation as $G(Q(z_t))$.
Since $Q$ is initialized as the identity, reconstruction initially favors $z_t \approx V_t$, which is intended to keep the factored representation close to the input space of the pretrained dynamics model that we fine-tune on it.
The reward term encourages the reward-relevant groups to carry the information needed to predict $r_t$.

\textbf{Dynamics on the Factored Representation.}
With the projections $f_k$ and the decoder $Q$ frozen, we fine-tune the pretrained dynamics model $f_\theta$ of Section~\ref{sec:background} to predict each factored representation from the earlier ones in its clip and the actions that led to them, $(z_{<t}, a_{<t}) \mapsto z_t$.
Since $z_t$ has the shape of $V_t$, this requires no architectural change: $f_\theta$ is trained with the same objective $\mathcal{L}_{\mathrm{dyn}}$, with $z_t$ in place of $V_t$ in Eqs.~\eqref{eq:bg_pred} and~\eqref{eq:shortcut}.
The reward head of $f_\theta$ reads the agent tokens and is therefore not restricted to the reward-relevant groups, unlike $h$.
The behavior-cloning head is trained with gradients stopped, so $\mathcal{L}_{\mathrm{bc}}$ does not update $f_\theta$, and each loss term is divided by a running estimate of its magnitude.
A predicted factored representation $\hat{z}_t$ is rendered as $G(Q(\hat{z}_t))$.
\section{Experiments}
\label{sec:experiments}
\begin{figure}[t]
  \centering
  \includegraphics[width=0.9\linewidth]{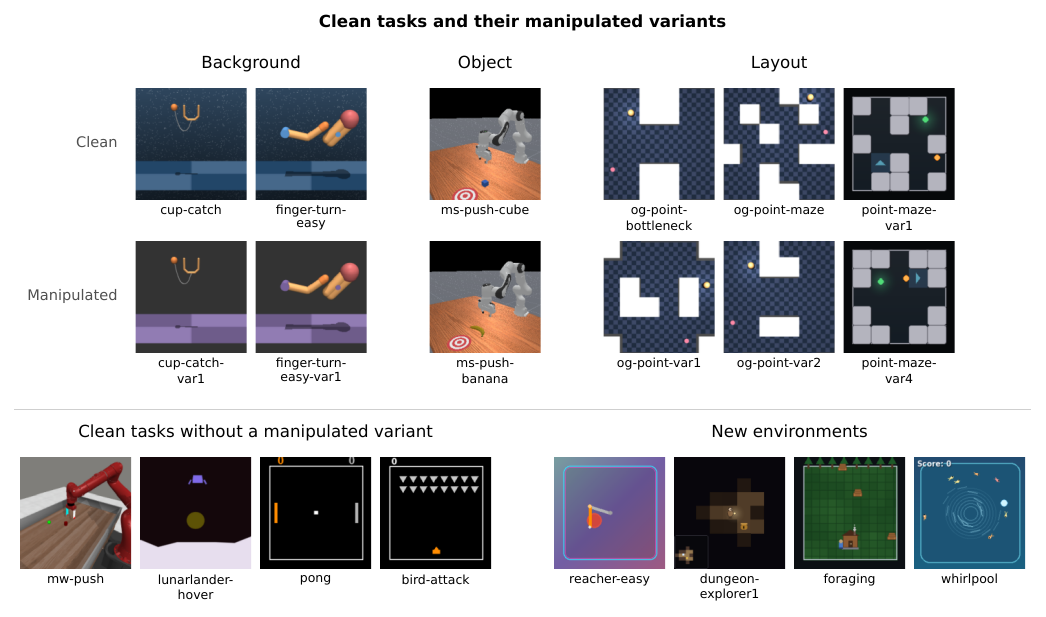}
  \caption{\textbf{Evaluation tasks.}
  \emph{Top:} six manipulated tasks, each a variant of a clean task (shown above it) that differs in the background (colors of the background and floor, with the same embodiment and goal), in the manipulated object, or in the maze layout.
 \emph{Bottom:} the four clean tasks without a manipulated variant, and four new environments with their own dynamics and rewards.
Clean tasks are part of the training corpus; manipulated tasks and new environments have no training data.
  Each image shows the first frame of an episode.}
  \label{fig:tasks}
\end{figure}
In this section, we evaluate whether fine-tuning a pretrained world model on the factored representation improves closed-loop planning, and whether the factored representation separates reward-irrelevant changes from its reward-relevant groups.
Our evaluation is structured to answer the following questions:
\begin{enumerate}
  \item Does planning with CausalDreamer reach higher task scores than the pretrained world model on clean tasks? (Section~\ref{sec:exp_clean})
  \item Does this difference persist, without fine-tuning, when one aspect of a clean task is changed, and does either model transfer to new environments? (Section~\ref{sec:exp_manip})
  \item Does the factored representation separate reward-irrelevant changes from its reward-relevant groups? (Section~\ref{sec:exp_ident})
\end{enumerate}
For each task we run one episode on each of 20 environment seeds, using the same seeds for every model, and we summarize results by the mean normalized score across tasks.

\subsection{Experimental Setup}
\label{sec:exp_setup}

\textbf{Models.}
All models share the frozen MMBench2 tokenizer, the training data and the planner, and differ only in the dynamics model:\\
\textit{CausalDreamer (ours):} the factored representation of Section~\ref{sec:method_causaldreamer} with dynamics fine-tuned from the pretrained checkpoint.\\
\textit{Pretrained:} the pretrained MMBench2 dynamics model~\citep{hansen2026mmbench2} from which CausalDreamer is fine-tuned, a 350M-parameter Dreamer 4-style world model.\\
\textit{Matched control:} the pretrained dynamics model fine-tuned for the same number of steps on the same data with the same objective, but on the tokenizer representation instead of the factored representation, which separates the effect of the factorization from the effect of additional training.

\textbf{Dataset.}
All models are trained on the MMBench2 pretraining corpus, which covers 200 tasks from 10 domains with 260 episodes per task of $224 \times 224$ frames labeled with ground-truth actions and rewards.
Its episodes are collected with three kinds of behavior policy: the expert policies of MMBench, mixed policies that perturb expert actions with action transforms, structured noise or random actions within an episode, and no-op actions; the logged action is always the action executed in the environment.
Actions are zero-padded to 16 dimensions with a per-task validity mask.
Although the corpus also provides low-dimensional states, all models use only RGB observations.

\textbf{Evaluation.}
We evaluate on twenty MMBench2 tasks (Figure~\ref{fig:tasks}), which we divide by their relation to the training data.\\
\textit{Clean tasks.}
The ten clean tasks are part of the training corpus: cup-catch and finger-turn-easy (DMControl), mw-push (Meta-World), ms-push-cube (ManiSkill), lunarlander-hover (Box2D), og-point-maze and og-point-bottleneck (OGBench), and pygame-point-maze-var1, pygame-pong and pygame-bird-attack (MiniArcade).\\
\textit{Manipulated tasks.}
Six tasks have no training data and are each obtained from a clean task by changing one aspect of it: the background, with the same embodiment and goal (cup-catch-var1 and finger-turn-easy-var1, from cup-catch and finger-turn-easy), the manipulated object (ms-push-banana, from ms-push-cube), or the maze layout (og-point-var1, og-point-var2 and pygame-point-maze-var4, from og-point-bottleneck, og-point-maze and pygame-point-maze-var1).\\
\textit{New environments.}
Four further tasks have no training data and no clean counterpart (pygame-reacher-easy, pygame-dungeon-explorer1, pygame-foraging and pygame-whirlpool).

All tasks are evaluated without fine-tuning, and each model receives only the text embedding of the task description.
We plan with the cross-entropy method at the MMBench2 settings, with a horizon of 32 steps, replanning every 16 steps, 256 samples, 32 elites and 4 iterations, and score each candidate by the sum of rewards predicted along its rollout.
The current frame is encoded together with up to 24 preceding real frames, matching the clip length used in training.
For each task we run one episode on each of 20 environment seeds, using the same seeds for all models and for a random policy that samples each action uniformly from $[-1, 1]$.
We report the normalized score $(R - R_{\text{random}})/(R_{\text{expert}} - R_{\text{random}})$, where $R$ is the episode return, $R_{\text{random}}$ is the mean return of the random policy on the same seeds, and $R_{\text{expert}}$ is the best return among the expert demonstrations of the task, so that the random policy scores $0$ and the expert $1$.
Since MMBench2 does not state the anchors of its normalized score, our scores are not directly comparable to the values it reports.

\textbf{Implementation Details.}
CausalDreamer and the matched control are trained on clips of $T = 24$ steps, sampled uniformly over all valid clip start positions of the corpus, so that tasks with more frames are sampled more often.
The projections $f_k$, the decoder $Q$ and the reward predictor $h$ are trained with $\mathcal{L}_{\mathrm{fac}}$, with $\lambda = 0.1$ in Eq.~\eqref{eq:fac}, for 30,000 optimizer steps of 256 clips (72 A100-40GB GPU-hours).
Each projection $f_k$ is a multilayer perceptron with three layers, and $Q$ is a residual multilayer perceptron, $Q(z) = z + q(z)$, whose last layer of $q$ is initialized to zero.
The dynamics model is fine-tuned for 30,000 optimizer steps of 256 clips, for CausalDreamer (140 A100-40GB GPU-hours) and for the matched control (143 A100-40GB GPU-hours).
The reward and behavior-cloning heads predict the rewards and actions of the next $M = 8$ steps, and the agent tokens are initialized from frozen CLIP text embeddings~\citep{radford2021learning} of each task's embodiment and instruction.

\subsection{Planning on Clean Tasks}
\label{sec:exp_clean}
\begin{table}[t]
  \centering
  \scalebox{0.72}{%
  \begin{tabular}{lrrrrrccc}
    \toprule
    & \multicolumn{5}{c}{Return} & \multicolumn{3}{c}{Normalized score} \\
    \cmidrule(lr){2-6} \cmidrule(lr){7-9}
    Task & Expert & Random & Pretrained & \begin{tabular}[b]{@{}c@{}}Matched\\control\end{tabular} & CausalDreamer & Pretrained & \begin{tabular}[b]{@{}c@{}}Matched\\control\end{tabular} & CausalDreamer \\
    \midrule
    cup-catch              & 1000.0 & 9.3 & 128.3 & 97.7 & 179.4 & 0.120 & 0.089 & \textbf{0.172} \\
    finger-turn-easy       & 1000.0 & 217.0 & 269.1 & 301.4 & 402.2 & 0.067 & 0.108 & \textbf{0.237} \\
    mw-push                & 1820.1 & 72.7 & 79.2 & 183.5 & 55.1 & 0.004 & \textbf{0.063} & $-0.010$ \\
    ms-push-cube           & 45.8 & 1.2 & 2.1 & 2.7 & 4.4 & 0.019 & 0.034 & \textbf{0.072} \\
    lunarlander-hover      & 469.2 & 58.6 & 84.3 & 53.9 & 73.7 & \textbf{0.063} & $-0.011$ & 0.037 \\
    og-point-maze          & 88.1 & $-96.4$ & 0.3 & $-25.4$ & $-22.9$ & \textbf{0.524} & 0.385 & 0.398 \\
    og-point-bottleneck    & 90.9 & $-79.5$ & $-3.3$ & $-11.2$ & 4.8 & 0.447 & 0.401 & \textbf{0.494} \\
    pygame-point-maze-var1 & 197.9 & 108.6 & 159.4 & 173.0 & 159.5 & 0.568 & \textbf{0.721} & 0.569 \\
    pygame-pong            & 4.0 & $-3.2$ & $-3.9$ & $-4.3$ & $-3.2$ & $-0.083$ & $-0.145$ & \textbf{0.007} \\
    pygame-bird-attack     & 110.0 & $-2.8$ & $-0.6$ & 0.0 & $-1.5$ & 0.020 & \textbf{0.025} & 0.012 \\
    \midrule
    Mean                   &        &         &        &        &         & 0.175 & 0.167 & \textbf{0.199} \\
    \bottomrule
  \end{tabular}%
  }
  \caption{\textbf{Clean tasks.} Mean episode return of the expert reference, a random policy, the pretrained model, the matched control and CausalDreamer, and the normalized score of each model (Section~\ref{sec:exp_setup}); higher is better. The best normalized score in each row is in bold.}
  \label{tab:clean}
\end{table}
\textbf{CausalDreamer plans better than the pretrained model on clean tasks.}
Over the ten clean tasks, CausalDreamer reaches a mean normalized score of $0.199$, against $0.175$ for the pretrained model (Table~\ref{tab:clean}).
The matched control, fine-tuned for the same number of steps on the tokenizer representation, reaches $0.167$ on its first 10 seeds, below the pretrained model, so additional fine-tuning alone does not account for the improvement of CausalDreamer.
It scores higher on six of the ten tasks, with the largest gains on finger-turn-easy ($0.237$ vs.\ $0.067$), pygame-pong ($0.007$ vs.\ $-0.083$), ms-push-cube ($0.072$ vs.\ $0.019$) and cup-catch ($0.172$ vs.\ $0.120$), and lower on four, most clearly on og-point-maze ($0.398$ vs.\ $0.524$).
CausalDreamer scores above the random policy on nine of the ten tasks; the exception is mw-push, where both models remain within $0.01$ of random actions.

\subsection{Planning on Manipulated Tasks and New Environments}
\label{sec:exp_manip}
\begin{table}[t]
  \centering
  \scalebox{0.85}{%
  \begin{tabular}{lrrrrcc}
    \toprule
    & \multicolumn{4}{c}{Return} & \multicolumn{2}{c}{Normalized score} \\
    \cmidrule(lr){2-5} \cmidrule(lr){6-7}
    Task & Expert & Random & Pretrained & CausalDreamer & Pretrained & CausalDreamer \\
    \midrule
    cup-catch-var1           & 1000.0 &   9.3   &  97.0   & 182.8   & 0.088          & \textbf{0.175} \\
    finger-turn-easy-var1    & 1000.0 & 217.0   & 284.8   & 464.6   & 0.087          & \textbf{0.316} \\
    ms-push-banana           &   45.8 &   0.9   &   1.1   &  10.9   & 0.005          & \textbf{0.222} \\
    og-point-var1            &   88.4 & $-97.7$ & $-49.4$ & $-53.0$ & \textbf{0.260} & 0.240 \\
    og-point-var2            &   90.7 & $-79.7$ & $-13.6$ & $-17.9$ & \textbf{0.388} & 0.363 \\
    pygame-point-maze-var4   &  198.6 & 118.4   & 170.3   & 160.7   & \textbf{0.647} & 0.527 \\
    \midrule
    Mean (manipulated)       &        &         &         &         & 0.246          & \textbf{0.307} \\
    \midrule
    pygame-reacher-easy      &  199.0 &   1.1   &  24.9   &   7.7   & \textbf{0.121} & 0.034 \\
    pygame-dungeon-explorer1 &  488.8 &  21.2   &  28.9   &  15.4   & \textbf{0.016} & $-0.012$ \\
    pygame-foraging          &   85.9 &  55.9   &  53.0   &  45.6   & \textbf{$-0.095$} & $-0.341$ \\
    pygame-whirlpool         &   41.5 & $-4.2$  & $-0.9$  &   0.4   & 0.072          & \textbf{0.101} \\
    \midrule
    Mean (new environments)  &        &         &         &         & \textbf{0.028} & $-0.055$ \\
    \bottomrule
  \end{tabular}%
  }
  \caption{\textbf{Manipulated tasks and new environments.} The first six tasks are manipulated variants of a clean task, changing its background, manipulated object or maze layout; the last four are set in new environments. Neither set of tasks has training data. Columns as in Table~\ref{tab:clean}.}
  \label{tab:manip}
\end{table}
\textbf{The improvement persists when a task is manipulated, most clearly under distractor backgrounds and a changed object.}
On the six manipulated tasks, CausalDreamer reaches a mean normalized score of $0.307$, against $0.246$ for the pretrained model (Table~\ref{tab:manip}).
Under a distractor background, it scores higher on both tasks (cup-catch-var1: $0.175$ vs.\ $0.088$; finger-turn-easy-var1: $0.316$ vs.\ $0.087$), and with a changed object it reaches $0.222$ on ms-push-banana, where the pretrained model stays at the level of random actions ($0.005$).
On all three of these tasks, the margin over the pretrained model is larger than on the corresponding clean task.
Under a changed maze layout, CausalDreamer scores lower on all three tasks, by $0.020$ to $0.120$.

\textbf{Neither model transfers to new environments.}
On the four tasks set in environments absent from training, neither model plans meaningfully better than a random policy (mean normalized scores of $0.028$ for the pretrained model and $-0.055$ for CausalDreamer), with the largest deficit on pygame-foraging, where random actions already obtain about two thirds of the expert return.
The improvement of CausalDreamer is therefore confined to the clean tasks and their manipulated variants.

\subsection{Analysis of the Factored Representation}
\label{sec:exp_ident}
\begin{figure}[t]
  \centering
    \begin{minipage}[t]{0.48\linewidth}
    \centering
    \includegraphics[height=1.45in]{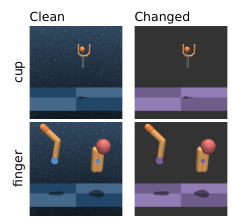}\\[1pt]
    {\small (a)}
  \end{minipage}\hfill
  \begin{minipage}[t]{0.48\linewidth}
    \centering
    \includegraphics[height=1.45in]{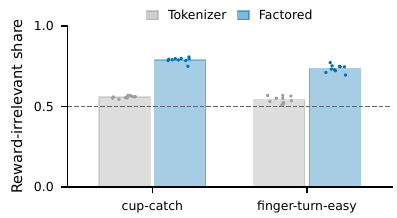}\\[1pt]
    {\small (b)}
  \end{minipage}
  \caption{\textbf{Effect of a background change on the representation.}
  (a)~Frames of cup-catch and finger-turn-easy rendered from the same simulator state with the clean background and with the background of the manipulated variant.
  (b)~Share of the change between the two renders that falls into the reward-irrelevant groups of the factored representation, and into the corresponding parts of the tokenizer representation; bars show the mean over 10 seeds, dots individual seeds, and the dashed line an equal split.}
  \label{fig:analysis}
\end{figure}
\textbf{The factored representation separates reward-irrelevant changes from its reward-relevant groups.}
The reward predictor reads only the reward-relevant groups $Z^{AR}$ and $Z^{\bar{A}R}$ (Eq.~\ref{eq:reward}), so content that does not affect the reward should be represented in the reward-irrelevant groups $Z^{A\bar{R}}$ and $Z^{\bar{A}\bar{R}}$ and leave the reward-relevant groups unchanged.
We test this by changing the background of a scene while keeping everything else fixed.
For cup-catch and finger-turn-easy, we render each clean task and its manipulated variant from the same initial state under the same actions, so that both episodes pass through identical simulator states and receive identical rewards and differ only in the background (Figure~\ref{fig:analysis}a).
We encode both episodes (10 seeds, 96 steps each) and measure which part of the representation changes, comparing the factored representation with the tokenizer representation $V_t$ split into four parts of the same size.
In the factored representation, 79\% (cup-catch) and 74\% (finger-turn-easy) of the change falls into the reward-irrelevant groups, so the reward-relevant groups receive only 21\% and 26\% (Figure~\ref{fig:analysis}b).
In the tokenizer representation, whose latent tokens each attend to the whole frame, the change is spread almost evenly, with 44\% and 46\% in the corresponding parts.

\section{Conclusion}
\label{sec:conclusion}

We presented CausalDreamer, which re-encodes the latent of the frozen tokenizer of a pretrained generative world model into a factored representation of four groups, organized by controllability and reward relevance, and fine-tunes the pretrained dynamics model to predict this representation.
In the factored representation, only the controllable groups receive the action and only the reward-relevant groups predict the reward, so this structure is added after pretraining, with the tokenizer frozen and the architecture and losses of the dynamics model unchanged.
On MMBench2, planning with CausalDreamer reaches a higher normalized score than the pretrained model on clean tasks seen during training and on manipulated variants of these tasks not seen during training, with the largest gains when the background or an object changes.
In new environments, neither model plans meaningfully better than a random policy, so the improvement does not extend to environments absent from training.
Our analysis shows that the factored representation separates reward-irrelevant changes from its reward-relevant groups: when only the background of a scene changes, most of the resulting change falls into the reward-irrelevant groups.

\clearpage

\bibliography{references}

\appendix

\section{Limitations and Future Work}
The factorization is imposed only when the factored representation is learned on top of the tokenizer: the dynamics model is fine-tuned with its original architecture, so every group, not only the controllable ones, attends to the action when it is predicted, and the reward head of the dynamics model reads all four groups rather than only the reward-relevant ones.
Making the dynamics model respect the same structure, for example with an attention mask that restricts the action and the reward head to the corresponding groups, is a direct next step.
Within the factored representation itself, the routing of the action is an architectural choice without an objective of its own, and we do not evaluate whether the controllable groups capture the effect of the action.
Our analysis covers changes that do not affect the reward, in two tasks, and does not test whether reward-relevant content is confined to the reward-relevant groups.
The objective encourages reward information to enter the reward-relevant groups but does not penalize it in the reward-irrelevant groups; adding such a term, for example an independence or mutual-information penalty as used by prior factorization methods~\citep{liu2023ifactor, wang2022denoised}, is a natural next step.
Finally, the gains vary across tasks: CausalDreamer scores lower than the pretrained model when the maze layout changes, and we report means over 20 seeds per task without significance tests.

\section{Related Work}
\label{sec:related_work}

\textbf{World models for control.}
A key challenge is scale, since a world model that plans across many tasks must predict high-dimensional video under many action spaces.
DreamerV3 learns a recurrent state-space model and trains a policy in imagination on over 150 tasks with a single configuration~\citep{hafner2023dreamerv3}, and TD-MPC2 plans in the latent space of a decoder-free model and trains one 317M-parameter agent on 80 tasks~\citep{hansen2024tdmpc2}.
Dreamer 4 scales this recipe to video by pairing a causal tokenizer, trained with masked autoencoding, with a block-causal transformer dynamics model trained with a shortcut forcing objective, and obtains diamonds in Minecraft from offline data alone~\citep{hafner2025dreamer4}.
MMBench2 trains a 350M-parameter Dreamer 4 world model on a 427-hour dataset of 210 tasks across ten domains, attributes its hallucinations mainly to gaps in data coverage, and evaluates it by model-predictive control with the cross-entropy method~\citep{hansen2026mmbench2}.
Our work keeps the MMBench2 model, data and planner unchanged and changes only the latent that its dynamics model predicts.

\textbf{Planning on frozen representations.}
A key challenge is that pixel prediction spends capacity on details a planner does not use, so recent world models predict in the feature space of a frozen encoder.
DINO-WM plans in the patch features of a frozen DINOv2 encoder~\citep{zhou2024dinowm}, and V-JEPA 2 plans with an action-conditioned predictor on top of a self-supervised video encoder~\citep{assran2025vjepa2}.
Causal-JEPA groups frozen DINOv2 features into object slots and masks whole object trajectories during training, so that the predictor must infer an object's state from the other objects~\citep{nam2026cjepa}.
It compares predictors that start from the same frozen embeddings on Push-T with CEM-based model-predictive control, and its object-level masking raises success from 60.67\% to 88.67\% over an object-centric baseline while using about 1\% of the latent features of the patch-based model.
Like Causal-JEPA, our work adds structure on top of a frozen encoder and measures its effect by closed-loop planning, but the structure comes from how the latent responds to actions and rewards rather than from objects.

\textbf{Factorized latent states.}
A key challenge is that reward-irrelevant and uncontrollable variation, such as a moving background, enters the latent state and distracts the policy, as benchmarks with distracting backgrounds show~\citep{stone2021distracting}.
Task Informed Abstractions train two models that jointly reconstruct the observation, one of which is adversarially kept from predicting reward~\citep{fu2021learning}, and deep bisimulation learns a representation in which states that yield the same rewards are mapped together~\citep{zhang2020learning}.
Iso-Dream separates controllable and non-controllable dynamics into two branches of a world model~\citep{pan2022iso}, and inverse-dynamics objectives identify the controllable part of the latent state from data~\citep{lamb2022guaranteed, islam2022agent}.
ASR characterizes the minimal state that is sufficient for control by the causal structure among states, actions and rewards~\citep{huang2022action}, Denoised MDPs sort information into four types by controllability and reward relevance and keep only the controllable, reward-relevant type~\citep{wang2022denoised}, and IFactor models the same four categories with causal influences between them and proves that they are identifiable in blocks~\citep{liu2023ifactor}.
Its theory also shows that all reward-relevant variables are needed for policy learning, including the uncontrollable ones that Denoised MDPs discard.
We adopt these four categories and route the action to the controllable ones, but we impose them on a pretrained tokenizer latent rather than learning a stochastic state from scratch.

\end{document}